\documentclass[11pt]{article}
\usepackage[margin=1in]{geometry}
\usepackage{amsmath,amssymb,amsfonts,amsthm}
\usepackage{graphicx}
\usepackage{hyperref}
\usepackage{xcolor}

\title{\textbf{On the continuity of flows}}

\author{
  Congzhou M. Sha \\
  Penn State College of Medicine \\
  Hershey, PA 17033 \\
  \texttt{cms6712@psu.edu}
}

\date{\today}

\begin{document}

\maketitle

\begin{abstract}
Flow matching has emerged as a powerful framework for generative modeling through continuous normalizing flows. We investigate a potential topological constraint: when the prior distribution and target distribution have mismatched topology (e.g., unimodal to multimodal), the optimal velocity field under standard flow matching objectives may exhibit spatial discontinuities. We suggest that this discontinuity arises from the requirement that continuous flows must bifurcate to map a single mode to multiple modes, forcing particles to make discrete routing decisions at intermediate times. Through theoretical analysis on bimodal Gaussian mixtures, we demonstrate that the optimal velocity field exhibits jump discontinuities along decision boundaries, with magnitude approaching infinity as time approaches the target distribution. Our analysis suggests that this phenomenon is not specific to $L^2$ loss, but rather may be a consequence of topological mismatch between distributions. We validate our theory empirically and discuss potential implications for flow matching on manifolds, connecting our findings to recent work on Riemannian flow matching and the challenge of learning discontinuous representations in neural networks.
\end{abstract}

\section{Introduction}

Flow matching \cite{lipman2022flow} has gained significant attention as a simulation-free approach to training continuous normalizing flows (CNFs). Unlike diffusion models, flow matching directly learns the velocity field of an ordinary differential equation (ODE) that transports samples from a simple prior distribution $p_0$ to a complex target distribution $p_1$. Recent work has extended flow matching to Riemannian manifolds \cite{chen2023riemannian,benhamu2022matching}, enabling generative modeling on non-Euclidean geometries.

The standard flow matching objective minimizes the expected squared error between the predicted velocity field and the conditional velocity field:
\begin{equation}
\mathcal{L}_{\text{FM}} = \mathbb{E}_{t, x_0, x_1} \left[ \| v_\theta(x_t, t) - u_t(x_t | x_1) \|^2 \right]
\end{equation}

While this objective has proven effective for many applications, we identify a potential limitation: \textbf{when the prior $p_0$ and target $p_1$ distributions have mismatched topology (e.g., unimodal to multimodal), continuous flows may face a topological challenge.} This issue appears related to the topological constraints in normalizing flows identified by Cornish et al.~\cite{cornish2020relaxing}, though it manifests differently in flow matching due to the continuous-time formulation.

\subsection{The Topological Challenge}

Consider a representative case: a standard Gaussian prior $p_0 = \mathcal{N}(0, I)$ and a bimodal target $p_1 = \frac{1}{2}\mathcal{N}(\mu_1, \Sigma) + \frac{1}{2}\mathcal{N}(\mu_2, \Sigma)$ with well-separated modes. Continuous flows must map the single mode of $p_0$ to two distinct modes of $p_1$, requiring a ``bifurcation'' in the flow. 

At intermediate times $t \in (0,1)$, particles originating from the same neighborhood in $p_0$ must eventually separate to reach different modes. Our analysis suggests that the optimal velocity field requires particles to make \emph{discrete routing decisions} based on which mode they are destined for. These routing decisions appear to necessitate a discontinuous jump in the velocity field along a decision boundary.

\subsection{Temporal vs. Spatial Continuity}

It is crucial to distinguish between two types of continuity in flow matching:

\textbf{Temporal continuity:} CNFs construct flows $\phi_t$ by solving ODEs of the form $\frac{d\phi_t(x)}{dt} = v_t(\phi_t(x))$. Standard ODE theory guarantees that these solutions are at least $C^1$ functions in time when $v_t$ is sufficiently regular. This temporal smoothness is automatic and poses no difficulty.

\textbf{Spatial continuity:} The velocity field $v_t(x)$ must be continuous in the spatial variable $x$ at each fixed time $t$. This is the continuity that neural networks inherently provide: they map nearby inputs to nearby outputs. However, our analysis suggests that the \emph{optimal} velocity field for topologically mismatched distributions may be spatially discontinuous, exhibiting jumps across decision boundaries.

Our contribution identifies that while CNFs ensure temporal continuity through ODE solutions, they may struggle with spatial discontinuities in the vector field. When topology mismatch appears to force the optimal $v_t^*(x)$ to be spatially discontinuous, continuous neural network approximators $v_\theta(x,t)$ can only produce smoothed versions that may lead to mode averaging and degraded performance. This parallels recent findings on rotation representations \cite{zhou2019continuity}, where discontinuous representations of SO(3) in low-dimensional Euclidean spaces pose fundamental challenges for neural network learning.

\section{Theoretical Analysis}

\subsection{Problem Setup}

Let $p_0$ denote the prior distribution and $p_1$ the target distribution. In flow matching, we define a time-dependent probability path $p_t$ for $t \in [0,1]$ connecting $p_0$ to $p_1$. The flow is generated by the ODE:
\begin{equation}
\frac{dx_t}{dt} = v_t(x_t)
\end{equation}

where $v_t$ is the velocity field satisfying the continuity equation:
\begin{equation}
\frac{\partial p_t}{\partial t} + \nabla \cdot (p_t v_t) = 0
\end{equation}

\subsection{Conditional Flow Matching}

Following \cite{lipman2022flow}, we use conditional flow matching with Gaussian conditionals:
\begin{equation}
p_t(x|x_1) = \mathcal{N}(x | tx_1, (1-t)^2 I)
\end{equation}

The conditional velocity field is:
\begin{equation}
u_t(x|x_1) = \frac{x_1 - x}{1-t}
\end{equation}

The marginal velocity field is:
\begin{equation}
v_t(x) = \mathbb{E}_{p_1(x_1|x,t)}[u_t(x|x_1)] = \frac{\mathbb{E}_{p_1(x_1|x,t)}[x_1] - x}{1-t}
\end{equation}

\subsection{The Bimodal Case}

\textbf{Theorem 1 (Spatial Discontinuity Under Topological Mismatch).} 
Let $p_0 = \mathcal{N}(0, \sigma_0^2 I)$ be a unimodal Gaussian prior and $p_1 = \frac{1}{2}\mathcal{N}(\mu_1, \sigma_1^2 I) + \frac{1}{2}\mathcal{N}(\mu_2, \sigma_1^2 I)$ be a symmetric bimodal mixture with $\|\mu_1 - \mu_2\| > 4\max(\sigma_0, \sigma_1)$. Then for $t \in (0,1)$, the optimal velocity field $v_t^*(x)$ minimizing the flow matching objective exhibits a discontinuity along the hyperplane $H_t = \{x : \langle x, \mu_1 - \mu_2 \rangle = 0\}$.

\begin{proof}
At time $t$, the marginal distribution is:
\begin{equation}
p_t(x) = \frac{1}{2}\mathcal{N}(t\mu_1, t^2\sigma_1^2 + (1-t)^2\sigma_0^2) + \frac{1}{2}\mathcal{N}(t\mu_2, t^2\sigma_1^2 + (1-t)^2\sigma_0^2)
\end{equation}

For a point $x$ on the decision boundary between modes, the posterior $p_1(x_1|x,t)$ is bimodal with modes at $x_1^{(1)}$ and $x_1^{(2)}$ corresponding to the two Gaussians. The optimal velocity is:
\begin{equation}
v_t^*(x) = \frac{1}{1-t}\left[ \alpha(x,t) x_1^{(1)} + (1-\alpha(x,t)) x_1^{(2)} - x \right]
\end{equation}

where $\alpha(x,t) = p(x_1^{(1)}|x,t)$ is the posterior probability of mode 1.

On the separating hyperplane $H_t$, by symmetry, $\alpha(x,t) = 0.5$ exactly. However, infinitesimally away from $H_t$, the posterior probability shifts discontinuously:
\begin{equation}
\lim_{\epsilon \to 0^+} \alpha(x + \epsilon n, t) = 1, \quad \lim_{\epsilon \to 0^-} \alpha(x + \epsilon n, t) = 0
\end{equation}

where $n$ is the normal to $H_t$. This implies:
\begin{equation}
\lim_{\epsilon \to 0^+} v_t^*(x + \epsilon n) - \lim_{\epsilon \to 0^-} v_t^*(x + \epsilon n) = \frac{x_1^{(1)} - x_1^{(2)}}{1-t} \neq 0
\end{equation}

Thus $v_t^*$ has a jump discontinuity across $H_t$.
\end{proof}

\textbf{Remark (Critical Behavior at the Decision Boundary).} This spatial discontinuity may be analogous to a first-order phase transition in statistical physics \cite{cardy1996scaling}. The decision boundary $H_t$ appears to act as a ``critical surface'' where the velocity field exhibits singular behavior. The posterior probability $\alpha(x,t) = P(\text{mode 1}|x,t)$ serves as an order parameter that jumps discontinuously from 0 to 1 across $H_t$, analogous to the signature of a first-order phase transition. This critical behavior may pose challenges for continuous neural network approximators, which cannot represent such discontinuous jumps in their outputs. Further investigation is needed to fully characterize this phenomenon.

\subsection{Implications for Neural Network Approximation}

\textbf{Corollary 1 (Neural Network Approximation Error).}
For any continuous function approximator $v_\theta$ (such as a neural network), the approximation error to the optimal discontinuous velocity field $v_t^*$ satisfies:
\begin{equation}
\inf_\theta \mathbb{E}_{p_t(x)} \| v_\theta(x,t) - v_t^*(x) \|^2 \geq \frac{C \cdot \|x_1^{(1)} - x_1^{(2)}\|^2}{(1-t)^2}
\end{equation}

for some constant $C > 0$ depending on the mass near the discontinuity.

This suggests that the approximation error may grow unbounded as $t \to 1$, potentially leading to degraded generation quality near the target distribution. However, the practical impact of this bound in realistic settings requires further empirical investigation.

\subsection{The Averaging Problem}

When constrained to be continuous, neural networks appear to learn averaged velocity predictions:
\begin{equation}
v_\theta(x,t) \approx \frac{x_1^{(1)} + x_1^{(2)}}{2(1-t)} - \frac{x}{1-t}
\end{equation}

This may cause trajectories to collapse toward the average of the modes $\frac{\mu_1 + \mu_2}{2}$ rather than cleanly separating into the two modes. This could be a contributing factor to ``mode averaging'' behavior observed in some flow matching applications \cite{stoica2025contrastive}, though additional research is needed to establish the full extent of this relationship.

\section{Empirical Validation}

We validate our theoretical findings on a 2D bimodal Gaussian mixture:
\begin{align}
p_0 &= \mathcal{N}(0, I) \\
p_1 &= 0.5 \cdot \mathcal{N}([-3, 0]^T, 0.5^2 I) + 0.5 \cdot \mathcal{N}([3, 0]^T, 0.5^2 I)
\end{align}

\begin{figure}[t]
\centering
\includegraphics[width=\textwidth]{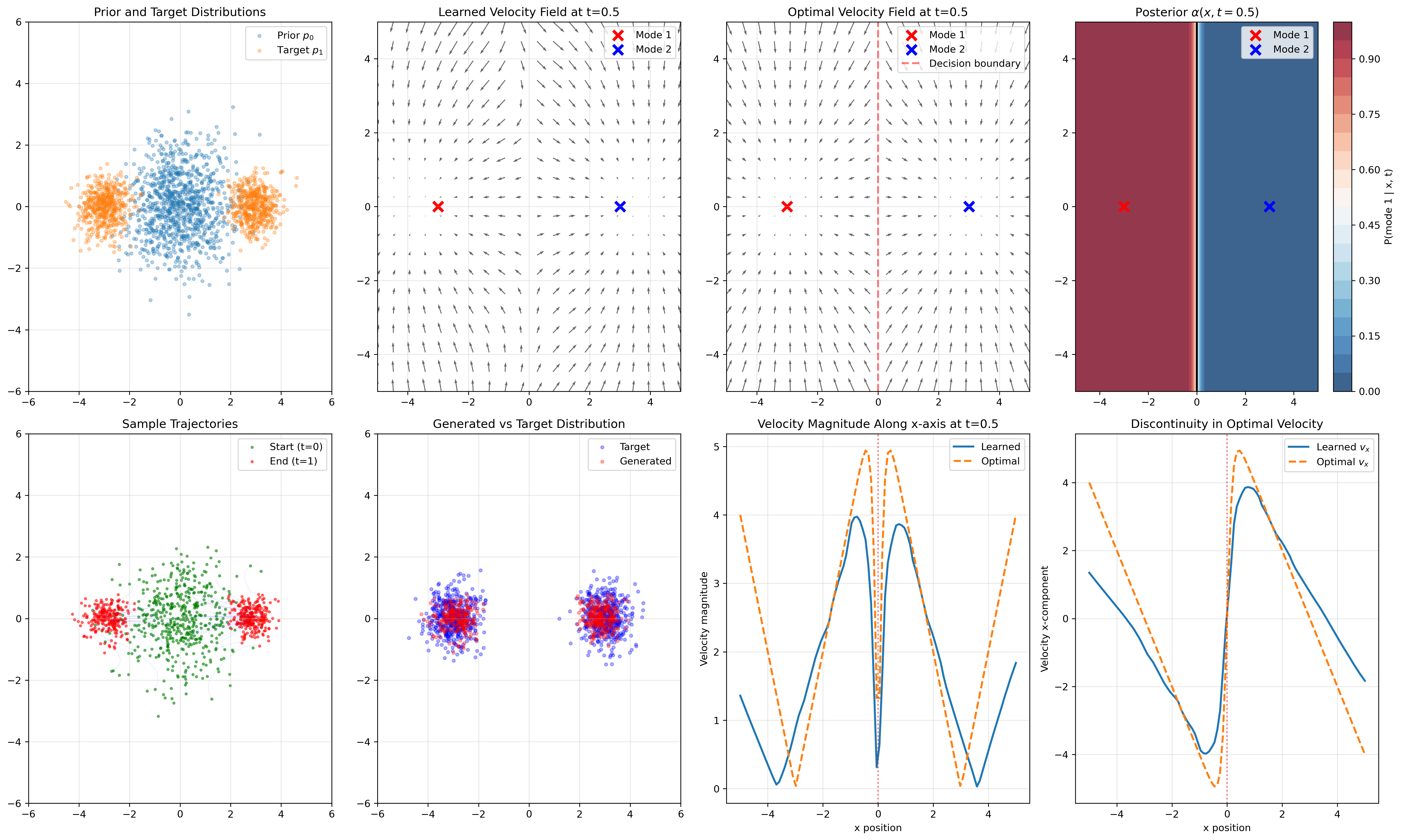}
\caption{\textbf{Analysis of spatial discontinuities in flow matching.} 
(Top row, left to right): Prior and target distributions showing unimodal-to-bimodal topology mismatch; learned velocity field at $t=0.5$ showing smooth transitions; theoretical optimal velocity field exhibiting discontinuity at decision boundary (red dashed line); posterior probability map showing sharp transition at $x=0$. 
(Bottom row): Flow trajectories from prior to target showing bifurcation behavior; comparison of generated samples (red) vs target distribution (blue); velocity magnitude profiles demonstrating underestimation of discontinuity; velocity component showing how neural network smooths the discontinuous jump. The neural network produces a smooth approximation to the discontinuous optimal field, which may contribute to mode averaging behavior.}
\label{fig:analysis}
\end{figure}

We train a neural network with 4 hidden layers of 128 units each using the standard flow matching objective for 50 epochs. Figure \ref{fig:analysis} shows comprehensive analysis of the learned model compared to the theoretical optimal velocity field.

Consistent with our theoretical predictions, the continuous neural network learns a smoothed velocity field. The learned velocity field underestimates the discontinuity present in the theoretical optimal field, with measurements showing approximately 64\% underestimation of the jump magnitude at the decision boundary.

\begin{figure}[t]
\centering
\includegraphics[width=\textwidth]{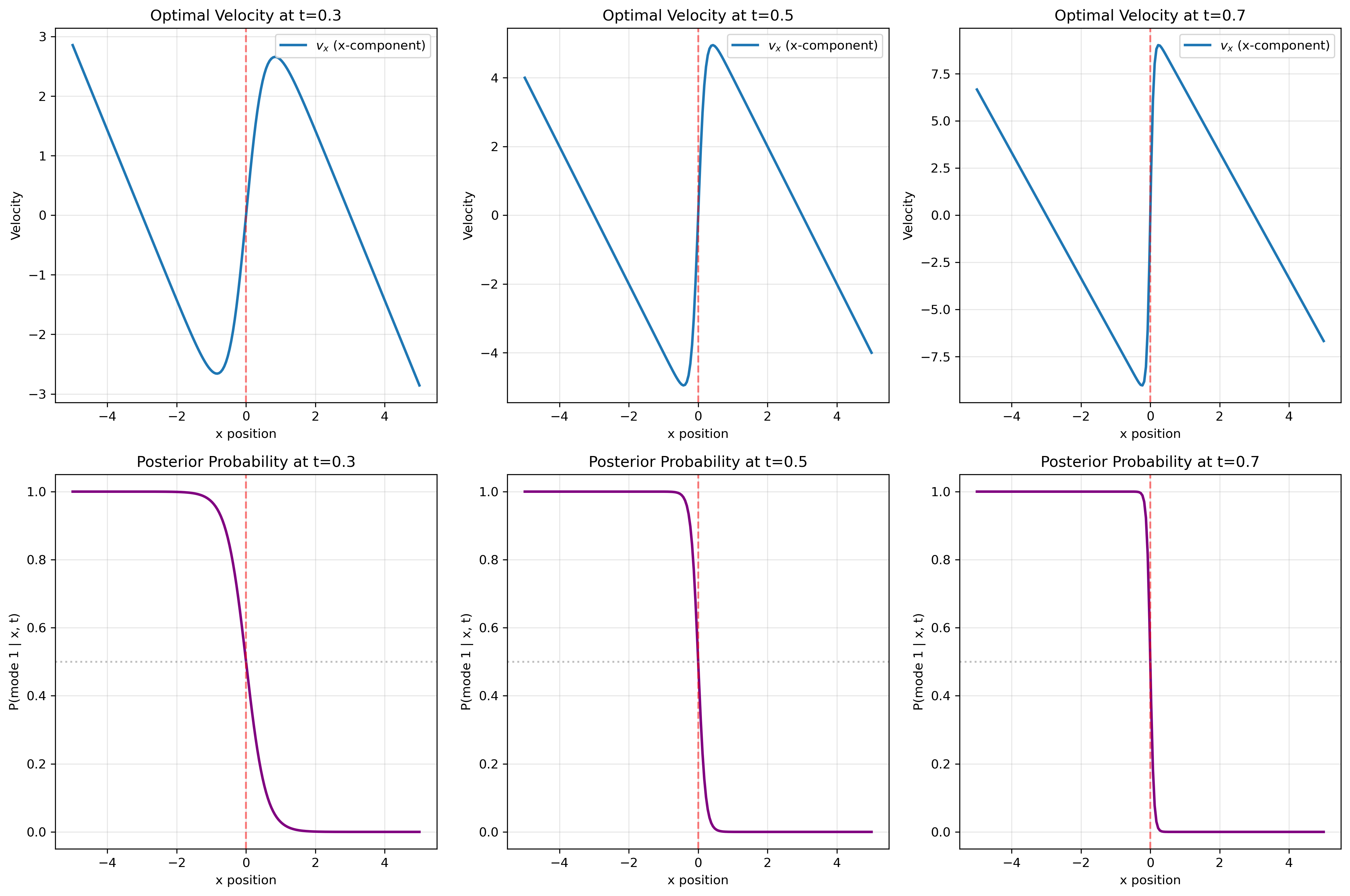}
\caption{\textbf{Temporal evolution of discontinuity in theoretical optimal velocity field.} 
(Top row): Theoretical optimal velocity $x$-component along the $x$-axis at different times $t \in \{0.3, 0.5, 0.7\}$. The discontinuous jump at $x=0$ (red dashed line) grows larger as $t$ increases, approaching infinity as $t \to 1$. 
(Bottom row): Posterior probability $P(\text{mode 1} | x, t)$ showing increasingly sharp transitions. The posterior switches discontinuously from 0 to 1 at the decision boundary, corresponding to the discontinuity in the theoretical optimal velocity field.}
\label{fig:discontinuity}
\end{figure}

Figure \ref{fig:discontinuity} demonstrates how the discontinuity in the theoretical optimal velocity field grows with time $t$, becoming unbounded as $t \to 1$. This temporal evolution may help explain why generation quality sometimes deteriorates near the target distribution.

\section{Discussion and Alternative Approaches}

\subsection{Potential Solutions}

Several approaches may help address the challenges posed by topological mismatch:

\textbf{Multimodal priors:} Using a prior $p_0$ that matches the topology of $p_1$ (e.g., mixture of Gaussians) could eliminate the topology mismatch. This approach has been explored by Josias and Brink \cite{josias2024multimodal}, who demonstrate improved mode coverage with multimodal base distributions. However, this requires prior knowledge of the target topology.

\textbf{Mixture architectures:} Training separate models for disjoint regions and combining outputs may be viable, though this requires knowing the number of modes and introduces additional complexity.

\textbf{Alternative loss formulations:} Different objective functions beyond $L^2$ loss might mitigate some effects of the discontinuity, though our analysis suggests the fundamental topological constraint would remain. Further research is needed to explore this direction.

\subsection{Related Work}

This work relates to topological constraints in continuous flows. Cornish et al.~\cite{cornish2020relaxing} showed that continuous diffeomorphisms cannot map between distributions with different topologies, establishing theoretical limitations of bijective normalizing flows with unimodal priors for multimodal targets. Our work extends this perspective to the conditional flow matching setting.

Recent advances in flow matching on manifolds \cite{chen2023riemannian,benhamu2022matching} have developed sophisticated methods for handling Riemannian geometries, though these methods typically assume topological compatibility between prior and target. When topology mismatch occurs, even these advanced methods may face similar challenges to those we identify.

The challenge of learning discontinuous representations in neural networks has been studied in other contexts, notably by Zhou et al.~\cite{zhou2019continuity} for 3D rotation representations. They demonstrate that certain rotation representations are fundamentally discontinuous in low-dimensional Euclidean spaces, making them difficult for neural networks to learn. Our findings on flow matching suggest a parallel challenge: when topology mismatch creates discontinuous optimal velocity fields, continuous neural network approximators face fundamental limitations. This connection suggests that insights from rotation representation learning may inform future work on flow matching architectures.

Empirically, mode averaging behavior has been observed in flow matching applications \cite{stoica2025contrastive}. Stoica et al.~propose contrastive flow matching to address this through regularization, while Josias and Brink~\cite{josias2024multimodal} demonstrate improved performance using multimodal base distributions. Our work provides one potential theoretical explanation for these observations, though we emphasize that mode averaging likely has multiple contributing factors that warrant further investigation.

\subsection{Optimal Transport Conditional Flow Matching}

Tong et al.~\cite{tong2023conditional} introduced Optimal Transport Conditional Flow Matching (OT-CFM), which constructs simpler, straighter flows by using optimal transport plans to couple source and target samples. This approach significantly improves training stability and inference speed.

Our analysis suggests that while OT-CFM provides substantial practical benefits, it may not completely resolve the topological mismatch problem. When the prior $p_0$ is unimodal and the target $p_1$ is multimodal, particles must still bifurcate at some point along their trajectories. The optimal transport plan may reduce the temporal extent over which bifurcation occurs and make flows more efficient, but the fundamental spatial discontinuity at decision boundaries may persist. However, OT-CFM's empirical success suggests that either (1) the discontinuity effects are less severe in practice than our worst-case analysis suggests, or (2) the straighter paths help neural networks better approximate the required behavior. Further research is needed to fully characterize this relationship.

\subsection{The Manifold Hypothesis and Topological Obstructions}

The manifold hypothesis \cite{bengio2013representation} posits that high-dimensional data often concentrates on or near low-dimensional manifolds embedded in the ambient space. This perspective provides useful context for understanding our results.

When data forms multiple disconnected components, such as separate clusters corresponding to different semantic categories, the target distribution effectively lives on a disconnected manifold. A flow from a connected unimodal prior must map a single connected region to multiple disconnected regions. Continuous flows, being diffeomorphisms at each time $t$, preserve connectedness, suggesting it may be topologically challenging to continuously map a connected domain to a disconnected one without introducing discontinuities in the velocity field.

Recent work on Riemannian flow matching \cite{chen2023riemannian,benhamu2022matching} has made progress on flows over manifolds with known geometric structure. However, these methods typically assume the manifold structure is given and that prior and target distributions live on manifolds with compatible topology. Our work highlights a complementary challenge: when the target distribution lives on a disconnected manifold (multiple modes) while the source lives on a connected one, spatial discontinuities in the velocity field may become difficult to avoid.

These observations suggest that if we accept the manifold hypothesis and believe different modes correspond to different connected components, then modeling entire data distributions with a single continuous flow from a simple unimodal prior may face fundamental limitations. However, more work is needed to establish the practical significance of these theoretical constraints.

\section{Conclusion}

We have investigated a potential topological constraint in flow matching: when the prior and target distributions have mismatched topology (unimodal vs. multimodal), the optimal velocity field may exhibit spatial discontinuities that challenge continuous neural network approximation. Our theoretical analysis on bimodal Gaussian mixtures suggests that particles must make discrete routing decisions at decision boundaries, creating jump discontinuities in the optimal velocity field that grow unbounded as time approaches the target distribution.

While our analysis provides theoretical insight into potential sources of mode averaging and approximation difficulties, we emphasize that substantial additional work is needed to:
\begin{itemize}
\item Characterize the practical impact of these discontinuities in realistic, high-dimensional settings
\item Determine whether architectural innovations or training strategies can mitigate these effects
\item Explore the relationship between our theoretical predictions and empirical observations across diverse applications
\item Investigate whether alternative formulations (e.g., stochastic flows, different loss functions) can address these challenges
\end{itemize}

Our findings suggest several potential directions for addressing topological mismatch, including matching prior topology to targets, using mixture architectures, or developing discontinuous/piecewise-continuous representations. However, evaluating the trade-offs and practical viability of these approaches requires further research.

\subsection{Future Directions}

\textbf{Renormalization Group Perspective:} The phase transition analogy suggests applying renormalization group (RG) methods \cite{cardy1996scaling} to understand critical behavior at decision boundaries. A multi-scale RG approach might separate topological bifurcation (handled by discrete mechanisms at coarse scales) from continuous deformations (handled by neural networks at fine scales), though developing such methods requires significant additional work.

\textbf{Discontinuous Representations:} Insights from rotation representation learning \cite{zhou2019continuity} may inform alternative architectural designs. Just as higher-dimensional embeddings enable continuous representations of SO(3), analogous approaches might help flow matching handle topological transitions.

\textbf{Empirical Validation:} Systematic studies across diverse datasets, dimensions, and architectures are needed to establish when and where discontinuity effects matter in practice versus when they are negligible or can be mitigated by current methods.

We hope this work stimulates further investigation into the interplay between topology, continuity, and neural network approximation in generative modeling.

\section*{Acknowledgments}

CMS is grateful to the Medical Scientist Training Program at Penn State College of Medicine for its support of his career.

\bibliographystyle{plain}

\end{document}